\documentclass[a4paper,twoside]{article}

\usepackage{epsfig}
\usepackage{subfigure}
\usepackage{calc}
\usepackage{amssymb}
\usepackage{amstext}
\usepackage{amsmath}
\usepackage{amsthm}
\usepackage{multicol}
\usepackage{pslatex}
\usepackage{apalike}

\usepackage[utf8]{inputenc}
\usepackage[english]{babel}
\usepackage{fancyhdr}

\usepackage{booktabs}

\usepackage{color}
\definecolor{darkblue}{rgb}{0.8,0.3,.5}

\definecolor{nullkommadreiachtvier}{rgb}{0.2,0.8,0.4}

\definecolor{thorstenfarbe}{rgb}{0.7,0.2,0.7}

\definecolor{dffarbe}{rgb}{0.6,0.4,0.3}

\definecolor{mzfarbe}{rgb}{0.3,0.4,0.8}

\usepackage{SCITEPRESS}     

\begin{document}

\title{Visual-Interactive Similarity Search for Complex Objects \\by Example of Soccer Player Analysis}

\author{\authorname{J\"{u}rgen Bernard\sup{1}, Christian Ritter\sup{1}, David Sessler\sup{1}, Matthias Zeppelzauer\sup{2}, J\"{o}rn Kohlhammer\sup{1,3}, and Dieter Fellner\sup{1,3}}
\affiliation{\sup{1}Technische Universit\"{a}t Darmstadt, Darmstadt, Germany}
\affiliation{\sup{2}St. P\"{o}lten University of Applied Sciences, St. P\"{o}lten, Austria}
\affiliation{\sup{3}Fraunhofer Institute for Computer Graphics Research, IGD, Darmstadt, Germany}
\email{\{juergen.bernard, christian.ritter, david.sessler, dieter.fellner\}@gris.tu-darmstadt.de, matthias.zeppelzauer@fhstp.ac.at, joern.kohlhammer@igd.fraunhofer.de}
}


\abstract{The definition of similarity is a key prerequisite when analyzing complex data types in data mining, information retrieval, or machine learning.
However, the meaningful definition is often hampered by the complexity of data objects and particularly by different notions of subjective similarity latent in targeted user groups.
Taking the example of soccer players, we present a visual-interactive system that learns users' mental models of similarity.
In a visual-interactive interface, users are able to label pairs of soccer players with respect to their subjective notion of similarity.
Our proposed similarity model automatically learns the respective concept of similarity using an active learning strategy.
A visual-interactive retrieval technique is provided to validate the model and to execute downstream retrieval tasks for soccer player analysis.
The applicability of the approach is demonstrated in different evaluation strategies, including usage scenarions and cross-validation tests.
}

\onecolumn \maketitle \normalsize \vfill


\begin{figure*}[t]
\vspace{-2mm}
\centering
\includegraphics[width=1.0\textwidth]{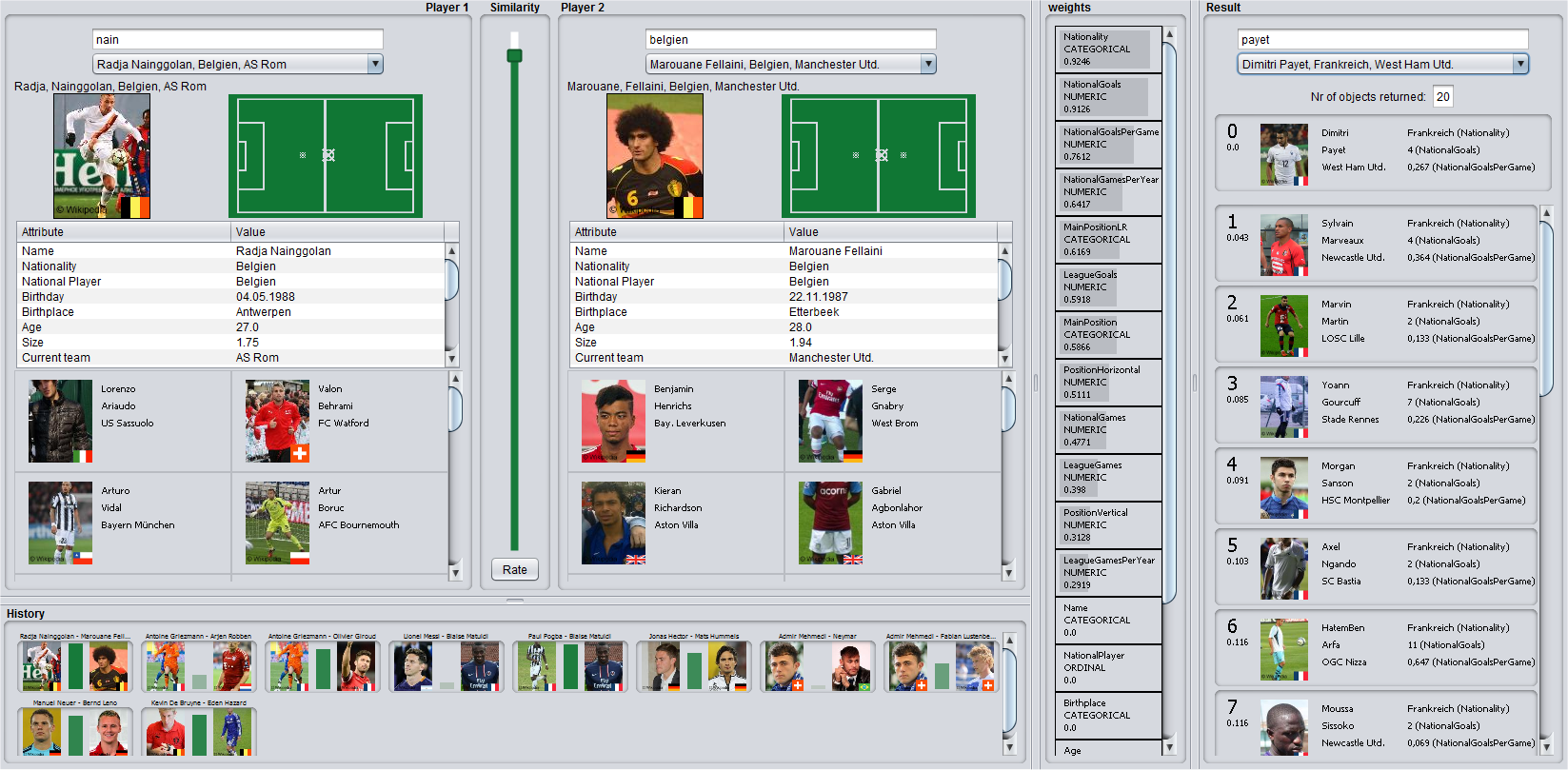}
\caption{Overview of the visual-interactive tool. Left: users are enabled to label the similarity between two soccer players (here: Radja Nainggolan and Marouane Fellaini, both from Belgium). The user's notion of similarity is propagated to the similarity learning model. Right: a visual search interface shows the model results (query: Dimitri Payet). The example resembles the similarity notion typically for a national trainer: only players from the same country have very high similarity scores. Subsequently, the nearest neighbors for Dimitri Payet are all coming from France and have a similar field position.}
\label{fig:titleImage}
\end{figure*}

\section{\uppercase{Introduction}}
\label{sec:introduction}

The way how similarity of data objects is defined and represented in an analytical system has a decisive influence on the results of the algorithmic workflow for downstream data analysis.
From an algorithmic perspective the notion of object similarity is often implemented with distance measures resembling an inverse relation to similarity.
Many data mining approaches necessarily require the definition of distance measures, e.g., for conducting clustering or dimension reduction.
In the same way, most information retrieval algorithms carry out indexing and retrieval tasks based on distance measures.
Finally, the performance of many supervised and unsupervised machine learning methods depends on meaningful definitions of object similarity.
The classical approach for the definition of object similarity includes the identification, extraction, and selection of relevant attributes (features), as well as the definition of a distance measure and optionally a mapping from distance to similarity. 
Furthermore, many real-world examples require additional steps in the algorithmic pipeline such as data cleansing or normalization.
In practice, quality measures such as precision and recall are used to assess the quality of the similarity models and the classifiers built upon them.


In this work, we strengthen the connection between the \emph{notion of similarity of individual users} and its adoption to the algorithmic definition of object similarity.
Taking the example of soccer players from European soccer leagues, a manager may want to identify previously unknown soccer players matching a reference player, e.g., to occupy an important position in the team lineup.
This is contrasted by a national coach who is also interested in selecting a good team. 
However, the national coach is independent from transfer fees and salaries while his choice is limited to players of the respective nationality.
The example sheds light on various remaining problems that many classical approaches are confronted with.
First, in many approaches designers do not know beforehand which definition of object similarity is most meaningful.
Second, many real-world approaches require multiple definitions of similarity for being usable for different users or user groups. 
Moreover, it is not even determined that the notion of similarity of single users remains constant. 
Third, the example of soccer players implicitly indicates that definition of similarity becomes considerably more difficult for high-dimensional data.
Finally, many real-word objects consist of mixed data, i.e. attributes of numerical, categorical, and binary type. However, most current approaches for similarity measurement are limited to numerical attributes.



We hypothesize that it is desirable to shift the definition of object similarity from an offline preprocessing routine to an integral part of future analysis systems.
In this way the individual notions of similarity of different users will be reflected more comprehensively.
The precondition for the effectiveness of such an approach is a means that enables users to communicate their notion of similarity to the system.
Logically, such a system requires the functionality to grasp and adopt the notion of similarity communicated by the user.
Provided that users are able to conduct various data analysis tasks relying on object similarity in a more dynamic and individual manner.
This requirement shifts the definition of similarity towards active learning approaches.
Active learning is a research field in the area of semi-supervised learning where machine learning models are trained with as few user feedback as possible, learning models that are as generalizable as possible. 
Beyond classical active learning, the research direction of this approach is towards visual-interactive learning allowing users to give feedback for those objects they have precise knowledge about.


We present a visual-interactive learning system that learns the similarity of complex data objects on the basis of user feedback.
The use case of soccer players will serve as a relevant and intuitive example.
Overall this paper makes three primary contributions.
\emph{First}, we present a visual-interactive interface that enables users to select two soccer players and to submit feedback regarding their subjective similarity. 
The set of labeled pairs of players is depicted in a history visualization for lookup and reuse.
\emph{Second}, a machine learning model accepts the pairwise notions of similarity and learns a similarity model for the entire data set.
An active learning model identifies player objects where user feedback would be most beneficial for the generalization of the learned model, and propagates them to a visual-interactive interface.
\emph{Third}, we present a visual-interactive retrieval interface enabling users to directly submit example soccer players to query for nearest neighbors.
The interface combines both validation support as well as a downstream application of model results.
The results of different types of evaluation techniques particularly assess the efficiency of the approach.
In many cases it takes only five labeled pairs of players to learn a robust and meaningful model.\\
The remaining paper is organized as follows.
Section \ref{sec:rw} shows related work.
We present our approach in Section \ref{sec:approach}.
The evaluation results are described in Section \ref{sec:usageScenario}, followed by a discussion in Section \ref{sec:discussion} and the conclusion in Section \ref{sec:conclusion}.


\section{\uppercase{Related Work}}
\label{sec:rw}

The contributions of this work are based on two core building blocks, i.e., visual-interactive interfaces (information visualization, visual analytics) and algorithmic similarity modeling (metric learning).
We provide a subsection of related work for both fields.

\subsection{Visual-Interactive Instance Labeling} 
\label{subsec:instanceLabeling}
We focus on visual-interactive interfaces allowing users to submit feedback about the underlying data collection.
In the terminology of the related work, a data element is often referred to as an \emph{instance}, the feedback for an instance is called a \emph{label}.
Different types of labels can be gathered to create some sort of learning model.
Before we survey existing approaches dealing with similarity in detail, we outline inspiring techniques supporting other types of labels.

Some techniques for learning similarity metrics are based on \emph{rules}.
The approach of \cite{FTKW08} allows users to create rules for ranking images based on their visual characteristics.
The rules are then used to improve a distance metric for image retrieval and categorization.
Another class of interfaces facilitates techniques related to \emph{interestingness} or relevance feedback strategies, e.g., to improve retrieval performance \cite{salton1997improving}.
One popular application field is evaluation, e.g,. to ask users which of a set of image candidates is best, with respect to a pre-defined quality criterion \cite{NW_DPID2016}.
In the visual analytics domain, relevance feedback and interestingness-based labeling has been applied to learn users' notions of interest, e.g., to improve the data analysis process.
Behrisch et al. \cite{BKSS14} present a technique for specifying features of interest in multiple views of multidimensional data.
With the user distinguishing relevant from irrelevant views, the system deduces the preferred views for further exploration.
Seebacher et al. \cite{SSJK16} apply a relevance feedback technique in the domain of patent retrieval, supporting user-based labeling of relevance scores.
Similar to our approach, the authors visualize the weight of different modalities (attributes/features).
The weights are subject to change with respect to the iterative nature of the learning approach. 
In the visual-interactive image retrieval domain the Pixolution Web interface\footnote{Pixolution, http://demo.pixolution.de, last accessed on September 22th, 2016} combines tag-based and example-based queries to adopt users' notions of interestingness. 
Recently the notion of interestingness was adopted to prostate cancer research.
A visual-interactive user interface enables physicians to give feedback about the well-being status of patients \cite{vahc2015}.
The underlying active-learning approach calculates the numerical learning function by means of a regression tree. 

\emph{Classification} tasks require categorical labels for the available instances.
Ware et al. \cite{ware2001interactive} present a visual interface enabling users to build classifiers in a visual-interactive way. The approach works well for few and well-known attributes, but requires labeled data sets for learning classifiers.
Seifert and Granitzer's \cite{SG10} approach outlines user-based selection and labeling of instances as meaningful extension of classical active learning strategies \cite{settles2009}. 
The authors point towards the potential of combining active learning strategies with information visualization which we adopt for both the representation of instances and learned model results.
H\"{o}ferlin et al. \cite{HNHWH12} define \emph{interactive} learning as an extension, which includes the direct manipulation of the classifier and the selection of instances. The application focus is on building ad-hoc classifiers for visual video analytics.
Heimerl at al. \cite{HKBE12} propose an active learning approach for learning classifiers for text documents. Their approach includes three user interfaces: basic active learning, visualization of instances along the classifier boundary, and interactive instance selection.
Similar to our approach the classification-based visual analytics tool by Janetzko et al. \cite{Janetzko2014} also applies to the soccer application domain. 
In contrast to our application goal, the approach supports building classifiers for interesting events in soccer games by taking user-defined training data into consideration.

User-defined labels for relevance feedback, interestingness, or class assignment share the idea to bind a single label to an instance, reflecting the classical machine learning approach ($f(i) = y$).
However, functions for learning the concept of similarity require a label representing the relation of pairs or groups of instances, e.g., in our case, $f(i_1, i_2) = y$, where y represents a similarity score in this case. 
Visual-interactive user interfaces supporting such learning functions have to deal with this additional complexity. 
A workaround strategy often applied for the validation of information retrieval results shows multiple candidates and asks the user for the most similar instances with respect to a given query. 
We neglect this approach since our users do not necessarily have the knowledge to give feedback for any query instance suggested by the system.
Rather, we follow a user-centered strategy where users themselves have an influence on the selection of pairs of instances.

Another way to avoid complex learning functions is allowing users to explicitly assign weights to the attributes of the data set \cite{ware2001interactive,jeong2009ipca}. 
The drawback of this strategy is the necessity of users knowing the attribute space in its entirety.
Especially when sophisticated descriptors are applied for the extraction of features (e.g., Fourier coefficients) or deep learned features, explicit weighting of individual features is inconceivable.
Rather, our approach applies an implicit attribute learning strategy.
While the similarity model indeed uses weighted attributes for calculating distances between instances (see Section \ref{rw:simiModeling}), an algorithmic model derives attribute weights based on the user feedback at object-level.
We conclude with a visual-interactive feedback interface where users are enabled to align small sets of instances on a two-dimensional arrangement \cite{bernardWSCG2014}.
The relative pairwise distances between the instances are then used by the similarity model.
We neglect strategies for arranging small sets of more than two instances in 2D since we explicitly want to include categorical and boolean attributes.
It has been shown that the interpretation of relative distances for categorical data is non-trivial \cite{vmv2014_ds}.

\subsection{Similarity Modeling}
\label{rw:simiModeling}



Aside from methods that employ visual interactive interfaces for learning the similarity between objects from user input as presented in the previous section, methods for the autonomous learning of similarity relations have been introduced~\cite{kulis2012metric,bellet2013survey}.
Human similarity perception is a psychologically complex process which is difficult to formalize and model mathematically. It has been shown previously that the human perception of similarity does not follow the rules of mathematical metrics, such as identity, symmetry and transitivity~\cite{tversky1977features}. Nevertheless, today most approaches employ distance metrics to approximate similarity estimates between two items (e.g., objects, images, etc.). 
Common distance metrics are Euclidean distance (L2 distance) and Manhattan distance (L1 distance) \cite{yu2008distance}, as well as warping or edit distance metrics.
The edit distance was, e.g., applied to the soccer domain in a search system where users can sketch trajectories of player movement \cite{vda2016}.

To better take human perception into account and to better adapt the distance metric to the underlying data and task an increasingly popular approach is to \emph{learn} similarity or distance measures from data (metric learning). For this purpose different strategies have been developed. 

In \emph{linear} metric learning the general idea is to adapt a given distance function (e.g., a Mahalanobis-like distance function) to the given task by estimating its parameters from underlying training data 
The learning is usually performed in a supervised or weakly-supervised fashion by providing ground truth in the form of (i) examples of similar and dissimilar items (positive and negative examples), (ii) continuous similarity assessments for pairs of items (e.g., provided by a human) and (iii) triplets of items with relative constraints, such as A is more similar to B and C \cite{bellet2013survey,xing2003distance}. During training the goal is to find parameters of the selected metric that maximizes the agreement between the distance estimates and the ground truth, i.e., by minimizing a loss function that measures the differences to the ground truth. The learned metric can then be used to better cluster the data or to improve the classification performance in supervised learning. 

Similar to these approaches, we also apply a linear model. 
Instead of learning the distance metric directly, we estimate the Pearson correlation between the attributes and the provided similarity assessments. 
In this way, the approach is applicable even to small sets of labeled pairs of instances.
The weights explicitly model the importance of each attribute and, as a by-product, enable the selection of the most important features for downstream approaches.
To facilitate the full potential, we apply weighted distance measures for internal similarity calculations, including measures for categorical \cite{boriah2008similarity} and boolean  \cite{cha2005enhancing} attributes.

\def \myVari {0.022}
\def \myHeight {0.03}

\begin{table*}[t]
\vspace{-2mm}
\small{

\begin{tabular}{p{0.135\linewidth}|p{0.31\linewidth}p{0.155\linewidth}p{0.155\linewidth}p{0.115\linewidth}}

	\toprule
	
	\textbf{Attribute} &
	\textbf{Description}&
	\textbf{Variable Type}&
	\textbf{Value Domain}&
	\textbf{Quality Issues}\\

	\midrule
	
	\textbf{Name}&
	\scriptsize{Name of the soccer player, unique identifier}&
	Nominal (String)&
	Alphabet of names&
	perfect
	
	\vspace{-2mm}
	
	\\

	\textbf{Description}&
	\scriptsize{Abstract of a player - for tooltips}&
	Nominal (String)&
	Full text &
	good	
	
	\vspace{-2mm}
	
	\\
	
	\textbf{Nationality}&
	\scriptsize{Nation of the player}&
	Nominal (String)&
	$103$ countries &
	perfect
	
	\vspace{-2mm}
	
	\\
	
	\textbf{National Team}&
	\scriptsize{National team, if applicable. Can be a youth team.}&
	Nominal (String)&
	$155$ nat. teams &
	sparse
	
	\vspace{-2mm}
	
	\\
	
	\textbf{Birthday}&
	\scriptsize{Day of birth}&
	(dd.mm.yyyy)&
	Date &
	perfect
	
	\vspace{-2mm}
	
	\\
	
	\textbf{Birthplace}&
	\scriptsize{Place of birth}&
	Nominal (String)&
	Alphabet of cities &
	good 
	
	\vspace{-2mm}
	
	\\
	
	\textbf{Size}&
	\scriptsize{Size of the player in meters}&
	Numerical &
	$[1.59, 2.03]$ &
	good
	
	\vspace{-2mm}
	
	\\
	
	\textbf{Current Team}&
	\scriptsize{Team of the player (end of last season)}&
	Nominal (String)&
	Alphabet of teams &
	perfect
	
	\vspace{-2mm}
	
	\\
	
	\textbf{Main Position}&
	\scriptsize{Main position on the field}&
	Nominal (String)&
	$13$ positions &
	perfect
	
	\vspace{-2mm}
	
	\\
	
	\textbf{Other positions}&
	\scriptsize{Other positions on the field}&
	Nominal (String)&
	$13$ positions &
	sparse, list
	
	\vspace{-2mm}
	
	\\
	
	\textbf{League Games}&
	\scriptsize{No. of games played in the current soccer league}&
	Numerical&
	$[0, 591]$ &
	sparse
	
	\vspace{-2mm}
	
	\\
	
	\textbf{League Goals}&
	\scriptsize{No. of goals scored in the current soccer league}&
	Numerical&
	$[0, 289]$ &
	sparse
	
	\vspace{-2mm}
	
	\\
	
	\textbf{Nat. Games}&
	\scriptsize{No. of games played for the current nat. team}&
	Numerical&
	$[0, 150]$ &
	sparse
	
	\vspace{-2mm}
	
	\\
	
	\textbf{Nat. Goals}&
	\scriptsize{No. of goals scored for the current nat. team}&
	Numerical&
	$[0, 71]$ &
	sparse
	
	\vspace{-2mm}
	
	\\
	
\bottomrule

\end{tabular}
}
\caption{Overview of primary data attributes about soccer players retrieved from DBpedia with SPARQL.}
\label{tab:primary}

\end{table*}


In \emph{non-linear} metric learning, one approach is to learn similarity (kernels) directly without explicitly selecting a distance metric. The advantage of kernel-based approaches is that non-linear distance relationships can be modeled more easily. 
For this purpose the data is first transformed by a non-linear kernel function. 
Subsequently, non-linear distance estimates can be realized by applying linear distance measurements in the transformed non-linear space \cite{abbasnejad2012survey,torresani2006large}. 
Other authors propose \emph{multiple kernel learning}, which is a parametric approach that tries to learn a combination of predefined base kernels for a given task \cite{gonen2011multiple}. 
Another group of non-linear approaches employs neural networks to learn a similarity function \cite{norouzi2012hamming,salakhutdinov2007learning}. 
This approach has gained increasing importance due to the recent success of deep learning architectures \cite{chopra2005learning,Zagoruyko_2015_CVPR,bell2015learning}. The major drawback of these methods is that they require huge amounts of labeled instances for training which is not available in our case. 

The above methods have in common that the learned metric is applied globally to all instances in the dataset. An alternative approach is \emph{local metric learning} that learns specific similarity measures for subsets of the data or even separate measures for each data item \cite{frome2007learning,weinberger2009distance,noh2010generative}. Such approaches have advantages especially when the underlying data has heterogeneous characteristics. A related approach are per-exemplar classifiers which even allow to select different features (attributes) and distance measures for each item. 
Per-exemplar classification has been applied successfully for different tasks in computer vision \cite{malisiewicz2011ensemble}. While our proposed approach to similarity modeling operates in a global manner, our active learning approach exploits local characteristics of the feature space by analyzing the density of labeled instances in different regions for making suggestions to the user. 


The approaches above mostly require large amounts of data as well as ground truth in terms of pairs or triplets of labeled instances. 
Furthermore, they rely on numerical data (or at least non-mixed data) as input. We propose an approach for metric learning for unlabeled data (without any ground truth) with mixed data types (categorial, binary, and numerical), which is also applicable to small datasets and data sets with  initially no labeled instances. 
For this purpose, we combine metric learning with active learning \cite{yang2012bayesian} and embed it in an interactive visualization system for immediate feedback.
Our approach allows the generation of useful distance metrics from a small number of user inputs. 

%
%


\section{\uppercase{Approach}}
\label{sec:approach}

\begin{figure*}[t]
\vspace{-2mm}
\centering
\includegraphics[width=1.0\textwidth]{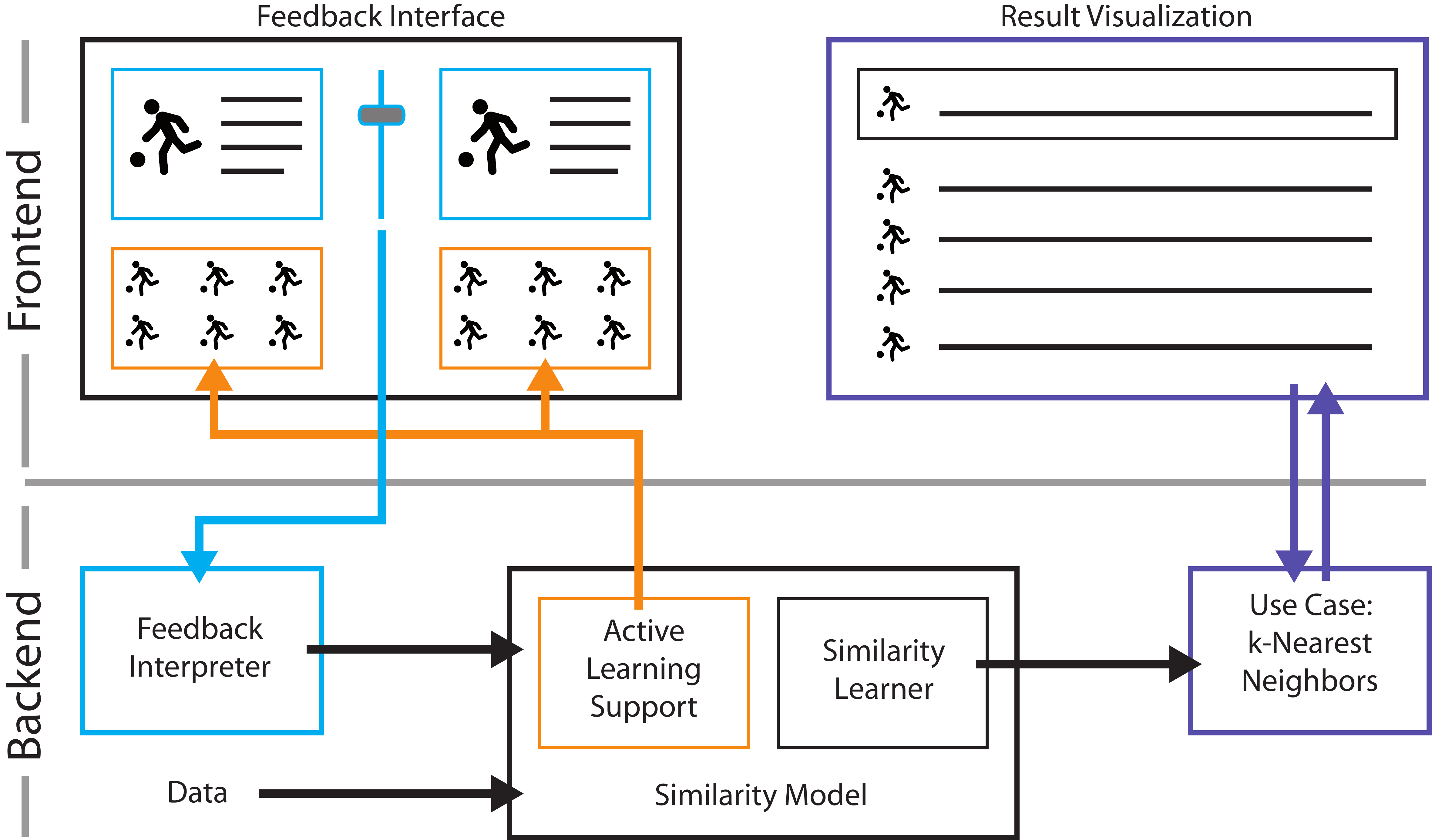}
\caption{Workflow of the approach. Users assign similarity scores for pairs of players in the feedback interface. In the backend, the feedback is interpreted (blue) and delegated to the similarity model. Active learning support suggests players to improve the model (orange). A kNN-search supports the use case of the workflow shown in the result visualization (purple).}
\label{fig:workflow}
\end{figure*}

An overview of the visual-interactive system is shown in Figure \ref{fig:titleImage}.
Figure \ref{fig:workflow} illustrates the interplay of the technical components assembled to a workflow.
In Sections \ref{approach:feedback}, \ref{approach:model}, and \ref{approach:result}, we describe the three core components in detail, after we discuss data characteristics and abstractions in Section \ref{sec:approach_data}.

\def \myVari {0.022}
\def \myHeight {0.03}

\begin{table*}[t]
\vspace{-2mm}
\small{

\begin{tabular}{p{0.19\linewidth}|p{0.31\linewidth}p{0.13\linewidth}p{0.16\linewidth}p{0.07\linewidth}}

	\toprule
	
	\textbf{Attribute} &
	\textbf{Description}&
	\textbf{Variable Type}&
	\textbf{Value Domain}&
	\textbf{Quality}\\

	\midrule
	
	\textbf{Age}&
	\scriptsize{Age of the player (end of last season)}&
	Numerical (int) &
	[16-43]&
	perfect
	
	\vspace{-2mm}
	
	\\
	
	\textbf{National Player}&
	\scriptsize{Whether the player has played as a national player}&
	Boolean&
	[false, true] &
	sparse
	
	\vspace{-2mm}
	
	\\
	
	\textbf{Nat. games p.a.}&
	\scriptsize{Average number of national games per year}&
	Numerical &
	$[0.0,24.0]$&
	sparse
	
	\vspace{-2mm}
	
	\\
	
	\textbf{Nat. goals per game}&
	\scriptsize{Average number of goals per national game}&
	Numerical &
	$< 3.0$ &
	sparse
	
	\vspace{-2mm}
	
	\\
	
	\textbf{League games p.a.}&
	\scriptsize{Average number of league games per year}&
	Numerical &
	$[0.0,73.5]$ &
	sparse
	
	\vspace{-2mm}
	
	\\
	
	\textbf{League goals p. game}&
	\scriptsize{Average number of goals per league game}&
	Numerical &
	$< 3.0$ &
	sparse
	
	\vspace{-2mm}
	
	\\
	
	\textbf{Position Vertical}&
	\scriptsize{The aggregated positions of the player as y-Coordinate (from ``Keeper'' to ``Striker'')}&
	Numerical &
	$[0.0, 1.0]$ &
	perfect
	
	\vspace{-2mm}
	
	\\
	
	\textbf{Position Horizontal}&
	\scriptsize{The aggregated positions of the player as x-Coordinate (from left to right)}&
	Numerical &
	$[0.0,1.0]$ &
	perfect
	
	\vspace{-2mm}
	
	\\
	
	\textbf{Main Position LR}&
	\scriptsize{The horizontal position of the player as String}&
	Nominal &
	\{left, center, right\} &
	perfect
	
	\vspace{-2mm}
	
	\\
	
\bottomrule

\end{tabular}
}
\caption{Overview of secondary data attributes about soccer players deduced from the primary data.}
\label{tab:secondary}

\end{table*}


\subsection{Data Characterization}
\label{sec:approach_data}

\subsubsection{Data Source}
Various web references provide data about soccer players with information differing in its scope and depth.
For example some websites offer information about market price values or sports betting statistics, while other sources provide statistics about pass accuracy in every detail.
Our prior requirement to the data is its public availability to guarantee the reproducibility of our experiments.
In addition, the information about players should be comprehensible for broad audiences and demonstrate the applicability. 
Finally, the attributes should be of mixed types (numerical, categorical, boolean). 
This is why ~Wikipedia\footnote{Wikipedia, https://en.wikipedia.org/wiki/Main\_Page, last accessed on September 22th, 2016} the free encyclopedia serves as our primary data source.
The structured information about players presented at Wikipedia is retrieved from ~DBpedia\footnote{DBpedia, http://wiki.dbpedia.org/about, last accessed on September 22th, 2016}. 
We access ~DBpedia with ~SPARQL \cite{sparql2008} the query language for RDF\footnote{RDF, https://www.w3.org/RDF, last accessed on September 22th, 2016} recommended by W3C\footnote{RDF, https://www.w3.org, last accessed on September 22th, 2016}.
We focused on the Europe's five top leagues (Premier League in England, Seria A in Italy, Ligue 1 in France, Bundesliga in Germany, and LaLiga in Spain).
Overall, we gathered 2,613 players engaged by the teams of respective leagues.

\subsubsection{Data Abstraction}
Table \ref{tab:primary} provides an overview of the available information about soccer players. 
Important attributes for the player (re-)identification are the unique name in combination with the nationality and the current team.
Moreover, a various numerical and categorical information is provided for similarity modeling.

Table \ref{tab:secondary} depicts the secondary data (i.e., attributes deduced from primary data).
Our strategy for the extraction of additional information is to obtain as much meaningful attributes as possible.
One benefit of our approach will be a weighing of all involved attributes, making the selection of relevant features for downstream analyses an easy task.
This strategy is inspired by user-centered design approaches in different application domains where we asked domain experts about the importance of attributes (features) for the similarity definition process \cite{BernardWKMSK13,bernardIJoDL2015}. 
One of the common responses was \emph{``everything can be important!''}
In the usage scenarios, we demonstrate how the similarity model will weight the importance of primary and secondary attributes with respect to the learned pairs of labeled players.

\subsubsection{Preprocessing}
One of the data-centered challenges was the sparsity of some data attributes.
This phenomenon can often be observed when querying less popular instances of concepts from DBpedia.
To tackle this challenge we removed attributes and instances from the data set containing only little information.
Remaining missing values were marked with missing value indicators, with respect to the type of attribute. 
For illustration purposes, we also removed players without an image in Wikipedia.
The final data set consists of 1,172 players. 
An important step in the preprocessing pipeline is normalization.
By default, we rescaled every numerical attribute into the relative value domain to foster metrical comparability.


\begin{figure*}[t]
\vspace{-2mm}
\centering
\includegraphics[width=1.0\linewidth]{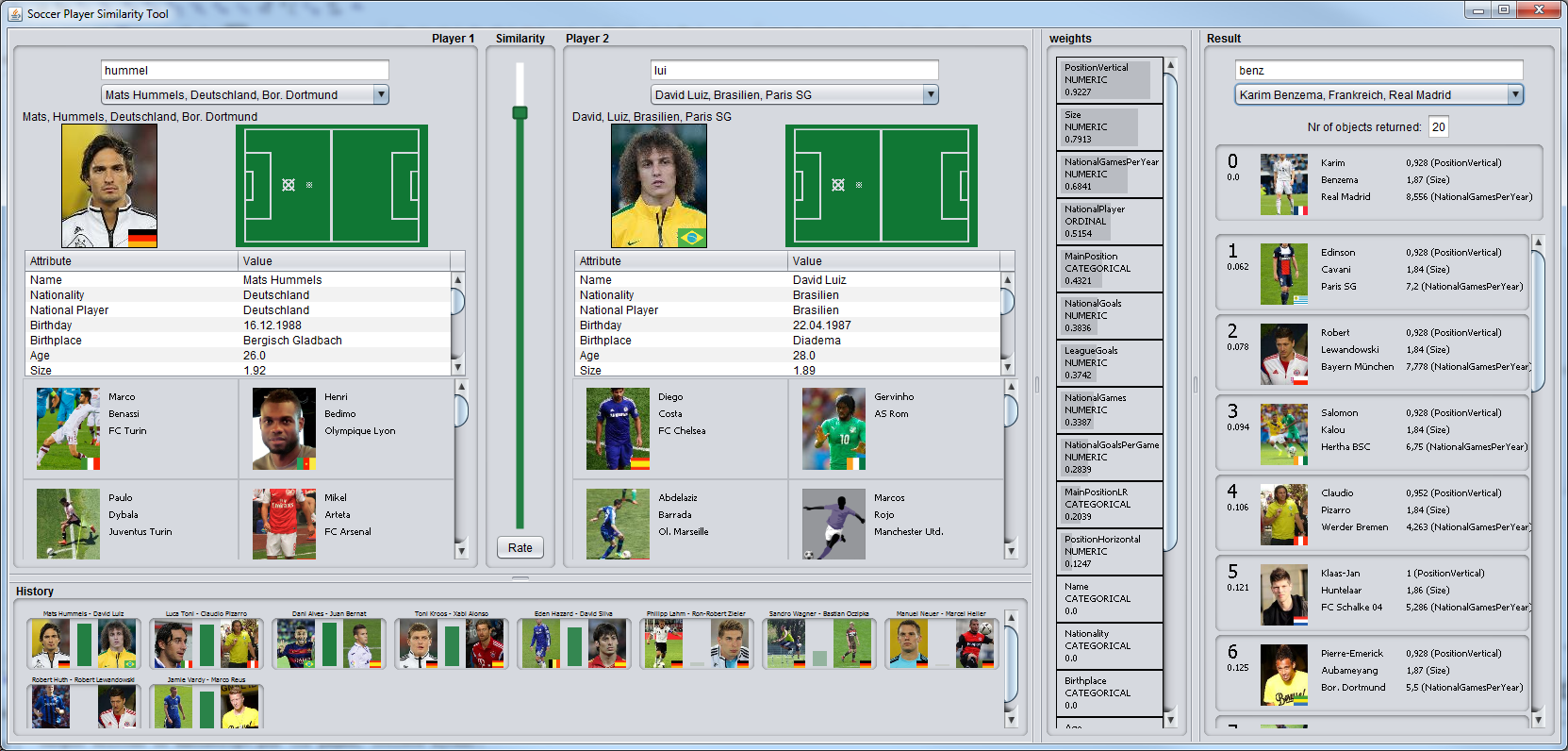}
\caption{Similarity model learned with stars in the European soccer scene. The history provides an overview of ten labeled pairs of players. The ranking of weighted attributes assigns high correlations to the vertical position, player size, and national games. Karim Benzema served as the query player: all retrieved players share quite similar attributes with Benzema.}
\label{fig:useCaseInternational}
\end{figure*}

\subsection{Visual-Interactive Learning Interface}
\label{approach:feedback}
One of the primary views of the approach enables users to give feedback about individual instances.
Examples of the feedback interface can be seen in the Figures \ref{fig:titleImage} and \ref{fig:useCaseInternational}.
The interface for the definition of similarity between soccer players shows two players in combination with a slider control in between.
The slider allows the communication of similarity scores between the two players.
We decided for a quasi-continuous slider, in accordance to the continuous numerical function to be learned.
However, one of possible design alternative would propose a feedback control with discrete levels of similarity.

Every player is represented with an image (when available and permitted), a flag icon showing the player's nationality, as well as textual labels for the player name and the current team.
These four attributes are also used for compact representations of players in other views of the tool.
The visual metaphor of a soccer field represents the players' main positions.
In addition, a list-based view provides the details about the players' attribute values.

The feedback interface combines three additional functionalities most relevant for the visual-interactive learning approach.\\
First, users need to be able to define and select players of interest.
This supports the idea to grasp detailed feedback about instances matching the users' expert knowledge \cite{SG10,HNHWH12}.
For this purpose a textual query interface is provided in combination with a combobox showing players matching a user-defined query.
In this way, we combine query-by-sketch and the query-by-example paradigm for the straightforward lookup of known players.\\
The second ingredient for an effective active learning approach is the propagation of instances to the user reducing the remaining model uncertainty. 
One crucial design decision determined that \emph{users should always be able to label players they actually know}.
Thus, we created a solution for the candidate selection combining automated suggestions by the model with the preference of users.
The feedback interface provides two sets of candidate players, one set is located at the left and the other one right of the interface.
Replacing the left feedback instance with one of the suggested players at the left will reduce the remaining entropy regarding the current instance at the right, and vice-versa.
However, we are aware that other strategies for proposing unlabeled instances exist.
Two of the obvious alternative strategies for labeling players would be a) providing a global pool of unlabeled players (e.g., in combination with drag-and-drop) or b) offering pairs of instances with low confidence.
While these two strategies may be implemented in alternative designs, we recall the design decision that users need to know the instances to be labeled.
In this way, we combine a classical active learning paradigm with the user-defined selection of players matching their expert knowledge.\\
Finally, the interface provides a history functionality for labeled pairs of players at the bottom of the feedback view (cf. Figure \ref{fig:titleImage}).
For every pair of players images are shown and the assigned similarity score is depicted in the center.




\subsection{Similarity Modeling}
\label{approach:model}

\subsubsection{Similarity Learning}

The visual-interactive learning interface provides feedback about the similarity of pairs of instances.
Thus, the feedback propagated to the system is according to the learning function $f(i_1, i_2) = y$ whereas $y$ is a numerical value between 0 (unsimilar) and 1 (very similar).
Similarity learning is designed as a two-step approach. First, every attribute (feature) of the data set is correlated with the user feedback. Second, pairwise distances are calculated for any given instance of the data set.

The correlation of attributes is estimated with Pearson's correlation coefficient.
Pairwise distances between categorical attributes are transformed into the numerical space with the Kronecker delta function. 
The correlation for a given attribute is then estimated between the labeled pairs of instances provided by the user and the distance in the value domain obtained by that attribute.
In the current state of the approach every attribute is correlated independently to reduce computation time and to maximize interpretability of the resulting weights. 
The result of this first step of the learning model is a weighting of the attributes that is proportional to the correlation.

In a second step, the learning model calculates distances between any pair of instances.
As the underlying data may consist of mixed attribute types, different distance measures are used for different types of attributes. 
For numerical data we employ the (weighted) Euclidean distance. 
For categorical attributes we choose the Goodall distance \cite{boriah2008similarity} since it is sensitive to the probability of attribute values and less frequent observations are assigned higher scores. 
The weighted Jaccard distance \cite{cha2005enhancing} is used for binary attributes.
The Jaccard distance neglects negative matches (both inputs false), which might be advantageous for many similarity concepts, i.e. the absence of an attribute in two items does not add to their similarity \cite{sneath1973numerical}.
After all distance measures have been computed in separate distance matrices all matrices are condensed into a single distance matrix by a weighted sum, where the weights represent the fraction of the sum of weights for each attribute type.


\subsubsection{Active Learning Strategy}

We follow an interactive learning strategy that allows for keeping the user in the loop.
To support the iterative nature, we designed an active learning strategy that fosters user input for instances for which no or little information is available yet.
As a starting point for active learning the user selects a known instance from the database.
Note that this is important as the user needs a certain amount of knowledge about the instance to make similarity assessments in the following (see Section \ref{approach:feedback}).
After an instance has been selected, we identify the attribute with the highest weight.
Next, we estimate the \emph{farthest} neighbors to the selected item under the given attribute for which no similarity assessments exist so far.
A set of respective candidates is then presented to the user.
This strategy is useful as it identifies pairs of items for which the system cannot make assumptions so far.
The user can now select one or more proposed items and add similarity assessments.
By adding assessments the coverage of the attribute space is successfully improved especially in sparse areas where little information was available so far.

\begin{figure}[t]
\centering
\includegraphics[width=1.0\columnwidth]{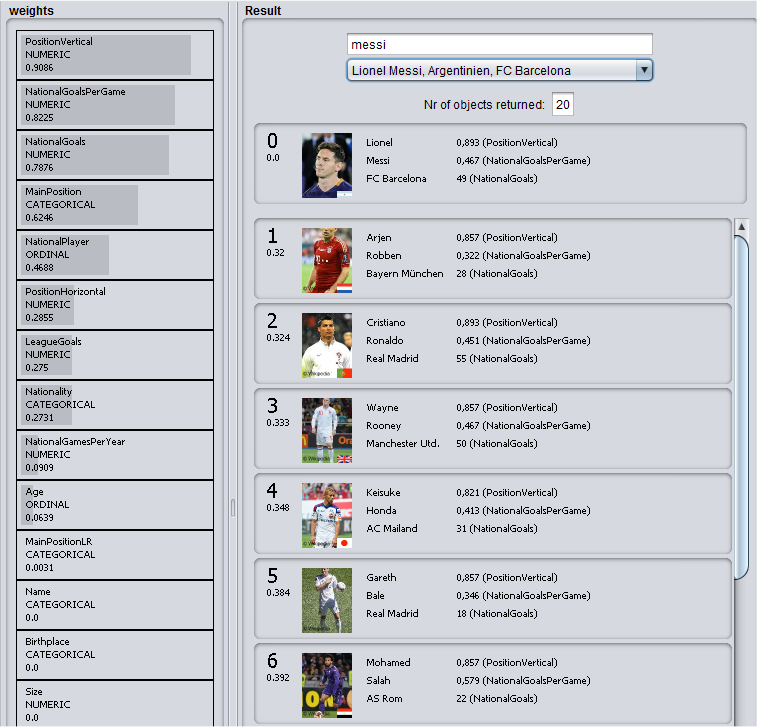}
\caption{Nearest neighbor search for Lionel Messi. Even if superstars are difficult to replace, the set of provided nearest neighbors is quite reasonable. Keisuke Honda may be surprising, nevertheless Honda has similar performance values in the national team of Japan.}
\label{fig:result}
\end{figure}

\subsubsection{Model Visualization}
Visualizing the \emph{output} of algorithmic models is crucial, e.g., to execute downstream analysis tasks (see Section \ref{approach:result}).
In addition, we visualize the \emph{current state of the model} itself.
In this way, designers and experienced users can keep track of the model improvement, its quality improvement, and its determinism.
The core black-box information of this two-step learning approach is the set of attribute weights representing the correlation between attributes and labeled pairs of instances.
Halfway right in the tool, we make the attribute weights explicit (between the feedback interface and the model result visualization), as it can be seen in the title figure.
An enlarged version of the model visualization is shown at the left of Figure \ref{fig:result} where the model is used to execute a kNN search for Lionel Messi.
From top to bottom the list of attributes is ranked in the order of their weights. 
It is a reasonable point of discussion whether the attribute weights should be visualized to the final group of users of such a system.
A positive argument (especially in this scientific context) is the transparency of the system which raises trust and allows the visual validation.
However, a counter argument is biasing users with information about the attribute/feature space.
Recalling that especially in complex feature spaces users do not necessarily know any attribute in detail, it may be a valid design decision to exclude the model visualization from the visual-interactive system.


\subsection{Result Visualization -- Visual-Interactive NN Search}
\label{approach:result}


We provide a visual-interactive interface for the visualization of the model output (see Figure \ref{fig:result}).
A popular use case regarding soccer players is the identification of similar players for a reference player, e.g., when a player is replaced in a team due to an upcoming transfer event.
Thus, the interface of the result visualization will provide a means to query for similar soccer players.
We combine a query interface (query-by-sketch, query-by-example) with a list-based visualization of retrieved players.
The retrieval algorithm is based on a standard k-NN search (k nearest neighbors) using the model output. 
For every list element of the result set a reduced visual representation of a soccer player is depicted, including the player's image, name, nationality, position, and team.
Moreover, we show information about three attributes contributing to the current similarity model significantly.
Finally, we depict rank information as well as the distance to the query for ever element.
The result visualization rounds up the functionality. 
Users can train individual similarity models of soccer player similarity and subsequently perform retrieval tasks. 
From a more technical perspective, the result visualization closes the feedback loop of the visual-interactive learning approach.
In this connection, users can analyze retrieved sets of players and give additional feedback for weakly learned instances.
An example can be seen in Figure \ref{fig:result} showing a retrieved result set for Lionel Messi used as an example query. 



\section{\uppercase{Evaluation}}
\label{sec:usageScenario}

\begin{figure}[t]
\centering
\includegraphics[width=1.0\columnwidth]{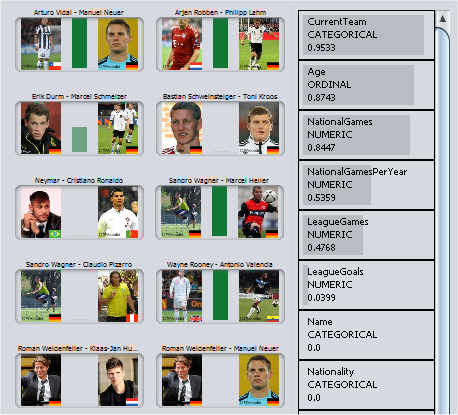}
\caption{Experiment with a mental model based on teams and player age. It can be seen that for ten labeled pairs of players the system is able to grasp this mental model.}
\label{fig:useCaseArtificialWeight}
\end{figure}

Providing scientific proof for this research effort is non-trivial since the number of comparable approaches is scarce.
Moreover, we address the challenge of dealing with data sets which are completely unlabeled at start, making classical quantitative evaluations with ground truth test data impossible.\\
In the following, we demonstrate and validate the applicability of the approach with different strategies.
In a first proof of concept scenario a similarity model is trained for an explicitly known mental model, answering the question whether the similarity model will be able to capture a human's notion of similarity.
Second, we assess the effectiveness of the approach in two usage scenarios. We demonstrate how the tool can be used to learn different similarity models, e.g., to replace a player in the team by a set of relevant candidates.
Finally, we report on experiments for the quantification of model efficiency.

\subsection{Proof of Concept - Fixed Mental Model}
\label{eval:1}

The first experiment assesses whether the similarity model of the system is able to grasp the mental similarity model of a user.
As an additional constraint, we limit the number of labeled pairs to ten, representing the requirement of very fast model learning.
As a proof of concept, we predefine a mental model and express it with ten labels. 
In particular, we simulate a fictitious user who is only interested in the age of players, as well as their current team.
In other words, a numerical and a categorical attribute defines the mental similarity model of the experiment.
If two players are likely identical with respect to these two attributes (age +-1) the user assigns the similarity score $1.0$.
If only one of the two attributes match, the user feedback is $0.5$ and if both attributes disagree a pair of players is labeled with the similarity score $0.0$.
The ten pairs of players used for the experiment are shown in Figure \ref{fig:useCaseArtificialWeight}.
In addition to the labeled pairs, the final attribute weights calculated by the system are depicted.
Three insights can be identified.
First, it becomes apparent that the two attributes with the highest weights exactly match the pre-defined mental model.
Second, the number of national games, the number of national games per year, and the number of league games also received weights. 
Third, the set of remaining attributes received zero weights.
While the first insight validates the experiment, the second insight sheds light on attributes correlated with the mental model.
As an example, we hypothesize that the age of players is correlated with the number of games.
This is a beneficial starting point for downstream feature selection tasks, e.g., when the model is to be implemented as a static similarity function.
Finally, the absence of weights for most other attributes demonstrates that only few labels are needed to obtain a precise focus on relevant attributes.

%

\subsection{Usage Scenario 1 - Top Leagues}
\label{eval:2}

The following usage scenario demonstrates the effectiveness of the approach.
A user with much experience in Europe's top leagues (Premier League, Seria A, Ligue 1, Bundesliga, LaLiga) rates ten pairs of prominent players with similarity values from very high to very low.
The state of the system after ten labeling iterations can be seen in Figure \ref{fig:useCaseInternational}.
The history view shows that high similarity scores are assigned to the pairs:
Mats Hummels vs. David Luiz, 
Luca Toni vs. Claudio Pizarro, 
Dani Alves vs. Juan Bernat,
Toni Kroos vs. Xabi Alonso, 
Eden Hazard vs. David Silva,
and Jamie Vardi vs. Marco Reus.
Comparatively low similarity values are assigned to the pairs 
Philipp Lahm vs. Ron-Robert Zieler, 
Sandro Wagner cs. Bastian Oczipka, 
Manuel Neuer vs. Marcel Heller, and 
Robert Huth vs. Robert Lewandowski.
The set of player instances resembles a vast spectrum of nationalities, ages, positions, as well as numbers of games and goals.
The resulting leaning model depicts high weights to the vertical position on the field, the player size, the national games per year.
Further, being a national player and the goals for the respective national teams contribute to the global notion of player similarity.
In the result visualization, the user chose Karim Benzema for the nearest neighbor search.
The result set (Edison Cavani, Robert Lewandowski, Salomon Kalou, Claudio Pizarro, Klass-Jan Huntelaar, Pierre-Emerick Aubameyang) represents the learned model quite well.
All these players are strikers, in a similar age, and very successful in their national teams.
In contrast, there is not a single player listed in the result who does not stick to the described notion of similarity.
In summary, this usage scenario demonstrates that the tool was able to reflect the notion of similarity of the user with a very low number of training instances.


%

\subsection{Usage Scenario 2 - National Trainer}
\label{eval:2b}
In this usage scenario, we envision to be a national trainer. 
Our goal is to engage a similarity model which especially resembles the quality of soccer players, but additionally takes the nationality of players into account.
As a result, similar players coming from the same country are classified similar. 
The reason is simple; players from foreign countries cannot be positioned in a national team.
Figure \ref{fig:titleImage} shows the history of the ten labeled pairs of players.
We assigned similar scores to two midfielders from Belgium, two strikers from France, two defenders from Germany, two goalkeepers from Germany, and two strikers from Belgium.
In addition, four pairs of players with different nationalities were assigned with considerably lower similarity scores, even if the players play at very similar positions.
The result visualization shows the search result for Dimitri Payet who was used as a query player.
In this usage scenario, the result can be used to investigate alternative players for the French national team.
With only ten labels, the algorithm retrieves exclusively players from France, all having expertise in the national team, and all sharing Dimitri Payets position (offensive midfielder).

\subsection{Quantification of Efficiency}
\label{eval:3}

\begin{figure*}[t]
\centering
\includegraphics[width=1.0\textwidth]{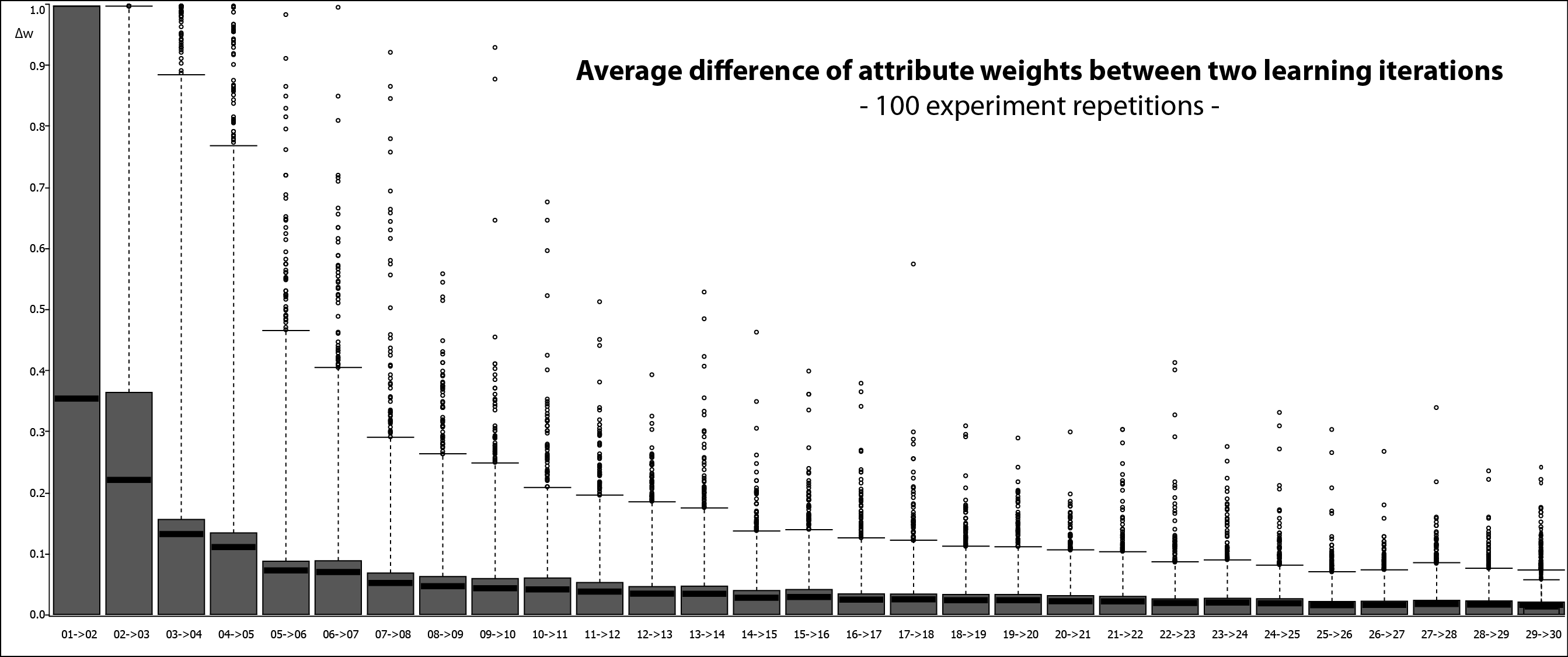}
\caption{Quantification of efficiency. The experiment shows differences in the attribute weighting between consecutive learning iterations. A saturation point can be identified, approximately after the 5th labeled pair of instances.}
\label{fig:weight_deltas_50}
\end{figure*}

In the final evaluation strategy, we conduct an experiment to yield quantitative results for the efficiency. 
We assess the `speed' of the convergence of the attribute weighting for a given mental model, i.e. how many learning iterations the model needs to achieve stable attribute weights.
The independent variable of the experiment is the number of learned iterations, i.e., the number of instances already learned by the similarity model.
The dependent variable is the change of the attribute weights of the similarity model between two learning iterations, assessed by the quantitative variable $\Delta w$.
To avoid other degrees of freedom, we fix the mental model used in the experiment. 
For this purpose, a small group of colleagues all having an interest in soccer  defined labels of similarity for 50 pairs of players.\\
To guarantee robustness and generalizability, we run the experiment 100 times.
Inspired by cross-validation, the set of training instances is permuted in every run.
The result is depicted in Figure \ref{fig:weight_deltas_50}.
Obviously, the most substantial difference of attribute weights is in the beginning of the learning process between the 1st and the 2nd learning iteration ($\Delta w = 0.36$).
In the following, the differences significantly decrease before reaching a saturation point approximately after the 5th iteration.
For the 6th and later learning iterations $\Delta w$ is already below $0.1$ and $0.03$ after the 30th iteration.
To summarize, the approach only requires very few labeled instances to produce a robust learning model.
This is particularly beneficial when users have very limited time, e.g., important experts in the respective application field.


\section{\uppercase{Discussion}}
\label{sec:discussion}

In the evaluation section, we demonstrated the applicability of the approach from different perspectives.
However, we want to shed light on aspects that allow alternative design decisions, or may be beneficial subjects to further investigation. 

\paragraph{Similarity vs. Distance}
Distance measures are usually applied to approximate similarity relationships.
This is also the case in our work.
We are, however, aware that metric distances can in general not be mapped directly to similarities, especially when the dimension of the data becomes high and the points in the feature space move far away from each other.
Finding suitable mappings between distance and similarity is a challenging topic that we will focus on in future research.


\paragraph{Active Learning Strategy}
The active learning support of this approach builds on the importance (weights) of attributes to suggest new learning instances to be queried.
Thus, we focus on a scalable solution that takes the current state of the model into account and binds suggestions to previously-labeled instances.
Alternative strategies may involve other intrinsic aspects of the data (attributes or instances) or the model itself.
For example statistical data analysis, distributions of value domains, or correlation tests could be considered.
Other active learning strategies may be inspired by classification approaches, i.e., models learning categorical label information.
Concrete classes of strategies involve uncertainty sampling or query by committee \cite{settles2009}.

\paragraph{Numerical vs. Categorical}
This research effort explicitly addressed a complex data object with mixed data, i.e., objects characterized by numerical, categorical, and boolean attributes.
This class of objects is widespread in the real-world, and we argue that it is worth to address this additional analytical complexity.
However, coping with mixed data can benefit from a more in-depth investigation at different steps of the algorithmic pipeline.


\paragraph{Usability}
We presented a technique that actually works but has not been throughoutly evaluated with users.
Will users be able to interact with the system?
We did cognitive walkthroughs and created the designs in a highly interactive manner. 
Still, the question arises whether domain experts will appreciate the tool and be able to work with it in an intuitve way.

\section{\uppercase{Conclusion}}
\label{sec:conclusion}

We presented a tool for the visual-interactive similarity search for complex data objects by example of soccer players.
The approach combines principles from active learning, information visualization, visual analytics, and interactive information retrieval.
An algorithmic workflow accepts labels for instances and creates a model reflecting the similarity expressed by the user.
Complex objects including numerical, categorical, and boolean attribute types can be included in the algorithmic workflow.
Visual-interactive interfaces ease the labeling process for users, depict the model state, and represent output of the similarity model.
The latter is implemented by means of an interactive information retrieval technique.
While the strategy to combine active learning with visual-interactive interfaces enabling users to label instances of interest is special, the application by example of soccer players is, to the best of our knowledge, unique.
Domain experts are enabled to express expert knowledge about similar players, and utilize learned models to retrieve similar soccer players.
We demonstrated that only very few labels are needed to train meaningful and robust similarity models, even if the data set was unlabeled at start.

Future work will include additional attributes about soccer players, e.g., market values or variables assessing the individual player performance.
In addition, it would be interesting to widen the scope and the strategy to other domains, e.g., in design study approaches.
Finally, the performance of individual parts of the algorithmic workflow may be tested against design alternatives in future experiments.

\bibliographystyle{apalike}
{\small
\bibliography{references}}

\vfill
\end{document}